\documentclass{article}

\usepackage{arxiv}

\usepackage[utf8]{inputenc} 
\usepackage[T1]{fontenc}    
\usepackage{hyperref}       
\usepackage{url}            
\usepackage{booktabs}       
\usepackage{amsfonts}       
\usepackage{nicefrac}       
\usepackage{microtype}      
\usepackage{lipsum}		
\usepackage{graphicx}
\usepackage[square,sort,comma,numbers]{natbib}
\usepackage{doi}
\usepackage{bm}

\title{Edge Direction-invariant Graph Neural Networks for Molecular Dipole Moments Prediction}

\author{ \href{https://orcid.org/0000-0000-0000-0000}{\includegraphics[scale=0.06]{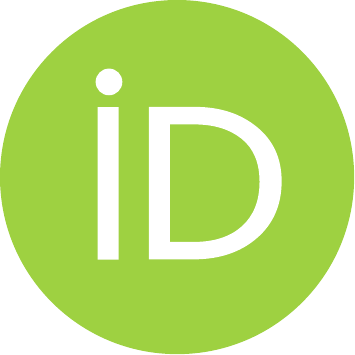}\hspace{1mm}Yang Jeong Park}\thanks{Use footnote for providing further
		information about author (webpage, alternative
		address)---\emph{not} for acknowledging funding agencies.} \\
	Department of Electrical and Computer Engineering\\
	Seoul National University\\
	08826 Seoul, Republic of Korea\\
	\texttt{yj3506@snu.ac.kr} \\
}

\renewcommand{\headeright}{}
\renewcommand{\undertitle}{}
\renewcommand{\shorttitle}{Edge Direction-invariant Graph Neural Networks for Molecular Dipole Moments Prediction}

\hypersetup{
pdftitle={Dipole moment article},
pdfsubject={q-bio.NC, q-bio.QM},
pdfauthor={Yang Jeong Park},
pdfkeywords={Dipole moment, Vectorial properties, Graph Neural Networks, Embedding equivariance},
}

\begin{document}
\maketitle

\begin{abstract}
	The dipole moment is a physical quantity indicating the polarity of a molecule and is determined by reflecting the electrical properties of constituent atoms and the geometric properties of the molecule. Most embeddings used to represent graph representations in traditional graph neural network methodologies treat molecules as topological graphs, creating a significant barrier to the goal of recognizing geometric information. Unlike existing embeddings dealing with equivariance, which have been proposed to handle the 3D structure of molecules properly, our proposed embeddings directly express the physical implications of the local contribution of dipole moments. We show that the developed model works reasonably even for molecules with extended geometries and captures more interatomic interaction information, significantly improving the prediction results with accuracy comparable to ab-initio calculations.
\end{abstract}


\section{Introduction}
The dipole moment has been considered a valuable property that can explain the chemical and physical behavior of molecules in various environments\cite{veit2020predicting}. It is of wide interest in applications such as the development of unknown materials\cite{pereira2018machine, gastegger2017machine, liu2020data, das2020dipole, matuszek2016defining, ioakimidis2008benchmarking, vo2019method}. It is closely related to the spectroscopic properties of a molecule\cite{liu2020data}; for example, infrared spectra of molecules can be obtained from dipole moment\cite{gastegger2017machine, nebgen2018transferable}. In particular, it is an important indicator that many compounds show a direct correlation between biological activity and dipole moment\cite{das2020dipole, matuszek2016defining, ioakimidis2008benchmarking}. Therefore, accurate and fast calculation of the dipole moment of arbitrary compounds sampled in a vast chemical space can accelerate recently attracting high-throughput screening (HTS) and have a powerful impact on material development and drug discovery. In HTS, quantum mechanics-based first-principle calculations that require a huge amount of computation, such as coupled cluster theory\cite{abe2018application, kongsted2002dipole} and density functional theory\cite{leenaerts2009water, hait2018accurate}, have been widely used.

Data-based machine learning approaches for quantum-mechanics properties computation have dramatically decreased the computational overhead required to predict various properties of molecules or crystals such as potential energy, dipole moments, and Gibbs free energy\cite{faber2017prediction, gastegger2017machine, pereira2018machine, faber2018alchemical, pinheiro2020machine}. However, modeling complex quantum mechanical effects between atoms using traditional machine learning techniques remains a major challenge. Recently, the MPNN framework\cite{gilmer2017neural}, which exchanges information between nodes through edges by expressing materials such as molecules and crystals as undirected graph structure data, has become the most widely used and successful model in this field\cite{gilmer2017neural, unke2019physnet, schutt2018schnet, klicpera2020directional, klicpera2020fast, choudhary2021atomistic, schutt2021equivariant}. In most of these kinds of operations, an atom is represented as a node of the molecular graph, connecting all atoms within a cutoff radius from the atom's position with an edge. The quantum mechanical interactions between atoms are appropriately modeled by message composed of atom features, which derived from the position coordinates and the atomic number of the atoms, and edge features developed from the interatomic distances or triplet angles. By exchanging generated messages across multiple layers to update node features and reading final node features, MPNN models have shown competitive performance in predicting various properties\cite{schutt2018schnet, unke2019physnet, klicpera2020directional}. These works of literature even provide efficient circumvention for calculating high-level properties such as solubility and protein docking\cite{lim2019predicting, fout2017protein, tang2020self}.

\begin{figure}
	\centering
	\includegraphics[width=300pt]{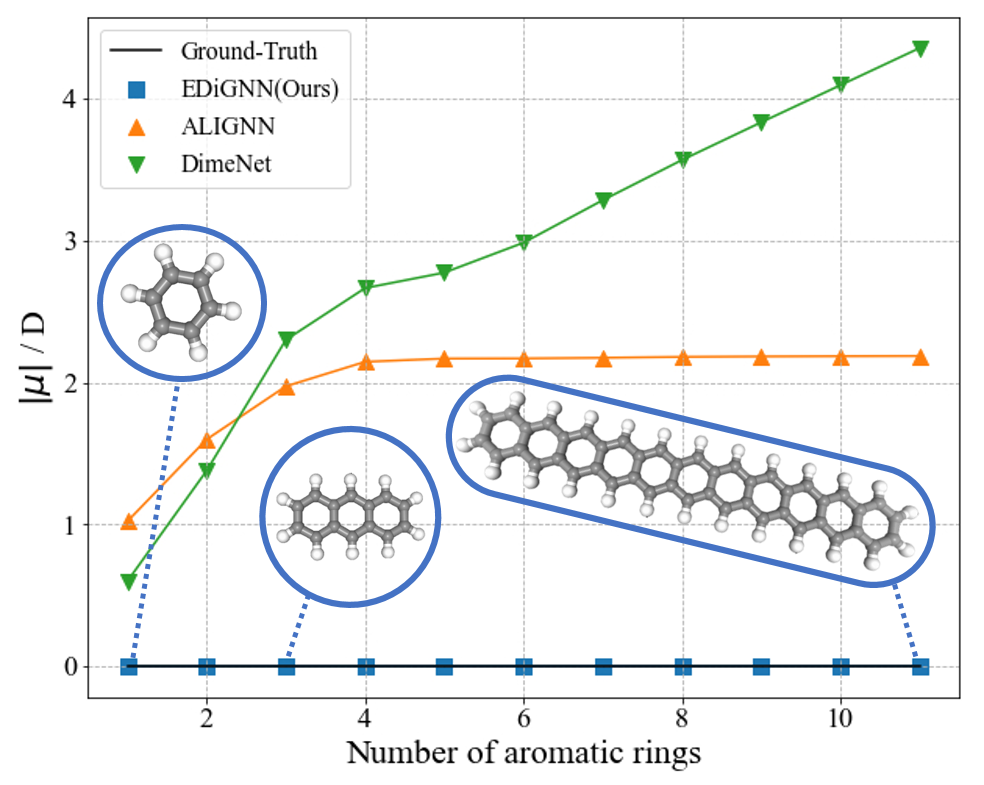}
	\caption{Comparison of dipole moment prediction results according to the number of benzene rings.}
	\label{fig:fig1}
\end{figure}

Despite these successes, it has been reported that MPNN-based models still have difficulties in predicting tensorial or vectorial properties\cite{schutt2021equivariant}. For example, the dipole moment is expressed as the sum of the partial charge and the vector component dependent on the position vector, so it greatly depends on the geometrical properties of the molecule. As a representative example, as shown in figure \ref{fig:fig1}, the polarization of molecules composed of benzene rings is canceled out by geometric symmetry, so the dipole moment should be predicted to be 0. However, even some modern models that employ methods of constructing messages from rotationally invariant interatomic distances fail to capture this geometric symmetry correctly. In the case of the model in which summation is selected as the readout function, the dipole moment increases as the number of benzene rings increases, and in the case of the model in which average is selected, the value deviating from 0 is continuously predicted. As an alternative to overcome these shortcomings, E(3) equivariant GNN using a rotational equivariant node representation has been recently proposed\cite{satorras2021n}, but it did not achieve dramatic improvement compared to the existing state-of-the-art model.

Our core idea to address these issues is to make the embeddings that model interatomic interactions invariant to edge direction. In a molecule represented as an undirected graph, pairs of atoms have edges in both directions. The existing rotational equivariant model does not guarantee that the representations of two edges are the same, which is unnatural to model the concept of dipole moment determined by partial charge and distance. Instead, our approach models the dipole moment by ensuring that one pair of atoms has one representation, and by predicting the dipole moment of the entire molecule as the sum of them, reflects the molecular geometry with high accuracy. Ablation studies have demonstrated the effectiveness of this direction-invariant representation. Our model approaches quantum-mechanical accuracy by significantly outperforming most modern MPNN models in predicting the published dipole moment dataset QM9.

\section{Related Works}
\subsection{Message passing neural networks for molecular graphs}
Graph neural networks (GNNs) have recently been popular in the field of materials science and engineering due to their advantages in accuracy.  In most applications, GNNs learn meaningful representations from a molecular graph $\mathcal{G}=(\mathcal{V}, \mathcal{E})$ consisting of nodes $\mathcal{V}$ representing atoms and edges $\mathcal{E}$ representing atom-atom pairs. In most previous work, an edge is considered to exist between two atoms if they have an interatomic distance within a fixed cutoff radius, even if there is no explicit bond between the atoms. Each node has a hidden state vector $\mathbf{h}_i^{l+1}\in\mathbb{R}^d$ representing the state of atom i at layer l, and each edge has a vector representation $\mathbf{e_{ij}}$ commonly referred to as an edge feature. Edge features are often embedded as a function of the interatomic distance $d_{ij}$. Message passing neural networks(MPNNs) are one of the most widely used frameworks for learning on graph. A general MPNN for embedded 3d molecular graphs can be written as 
\begin{equation}
    \mathbf{m}_i^{l+1} = \sum_{j\in \mathcal{N}(i)} M^l(\mathbf{h}_i^l, \mathbf{h}_j^l, \mathbf{e}_{ij})
\end{equation}
\begin{equation}
    \mathbf{h}_i^{l+1} = U^l(\mathbf{h}_i^l, \mathbf{m}_i^{l+1} )
\end{equation}
where $M^l$ and $U^l$ are an arbitrary message function and update function \cite{gilmer2017neural}. Information from atoms inside the receptive field, which means near than cutoff, is aggregated on a central node state $\mathbf{h}_i^L$ at the final layer $L$ of the network. The readout step computes the target from the node features of the whole graph using the readout function R, which is invariant to permutations.
\begin{equation}
    \hat{y} = R(\left\{\mathbf{h}_i^L | i \in \mathcal{G}\right\})
\end{equation}
It becomes clear that the representation of the central node is primarily local, but can be extended to the receptive field of the surrounding atoms through multiple layers to obtain an improvement in the expressive power. This allows for the possibility of MPNNs capturing long-range interactions, which can explain multi-body correlations of complex molecules. It was recently discovered that the expressive power of MPNNs is limited to the 1-Weisfeiler-Lehman isomorphism test \cite{xu2018powerful}, a classical method that attempts to determine whether two graphs are isomorphic via iterative color enhancement.

\subsection{Equivariant GNNs}
Most molecular properties such as potential energy or bandgap must satisfy invariance in transformations for the Euclidean group $E(3)$. Most existing MPNNs guarantee the invariance of their predictied properties by acting only on invariant inputs. However, in the case of vector properties such as dipole moments, equivariance should be satisfied instead of invariance. More recently, a class of models known as equivariant GNNs \cite{satorras2021n, batzner20223} have been developed that construct equivariant node and edge features by directly inputting equivariant geometric information, such as displacement vectors. A function between $\phi:X\rightarrow Y$ is equivariant to a group $G$ if 
\begin{equation}
    \phi(T_g(x)) = S_g(\phi(x)) \quad \forall g \in G, \forall x \in X
\end{equation}
where $T_g$ is the transformation on input space $X$ for the abstract group $g$ and $S_g$ is the transformation on output space $Y$\cite{satorras2021n}. When $S_g$ is an identity operator on $Y$, the function $\phi$ is invariant and the output is unchanged by the operator $T_g$. 
The rotational invariance of the hidden representation can be guaranteed by choosing the input of the message function and the update function as an interatomic distance defined as a scalar. In contrast, hidden representations for predicting vector quantities, such as atomic forces or dipole moments, must be transformed accordingly when the atomic geometry is transformed. The proposed method fixes the output of the message function as a transform-invariant scalar. Transform equivariance can be achieved by multiplying the scalar output by the coordinate embedding. During the entire update process, node and edge functions are separated to remain transform-invariant.

\begin{figure}
	\centering
	\includegraphics[width=300pt]{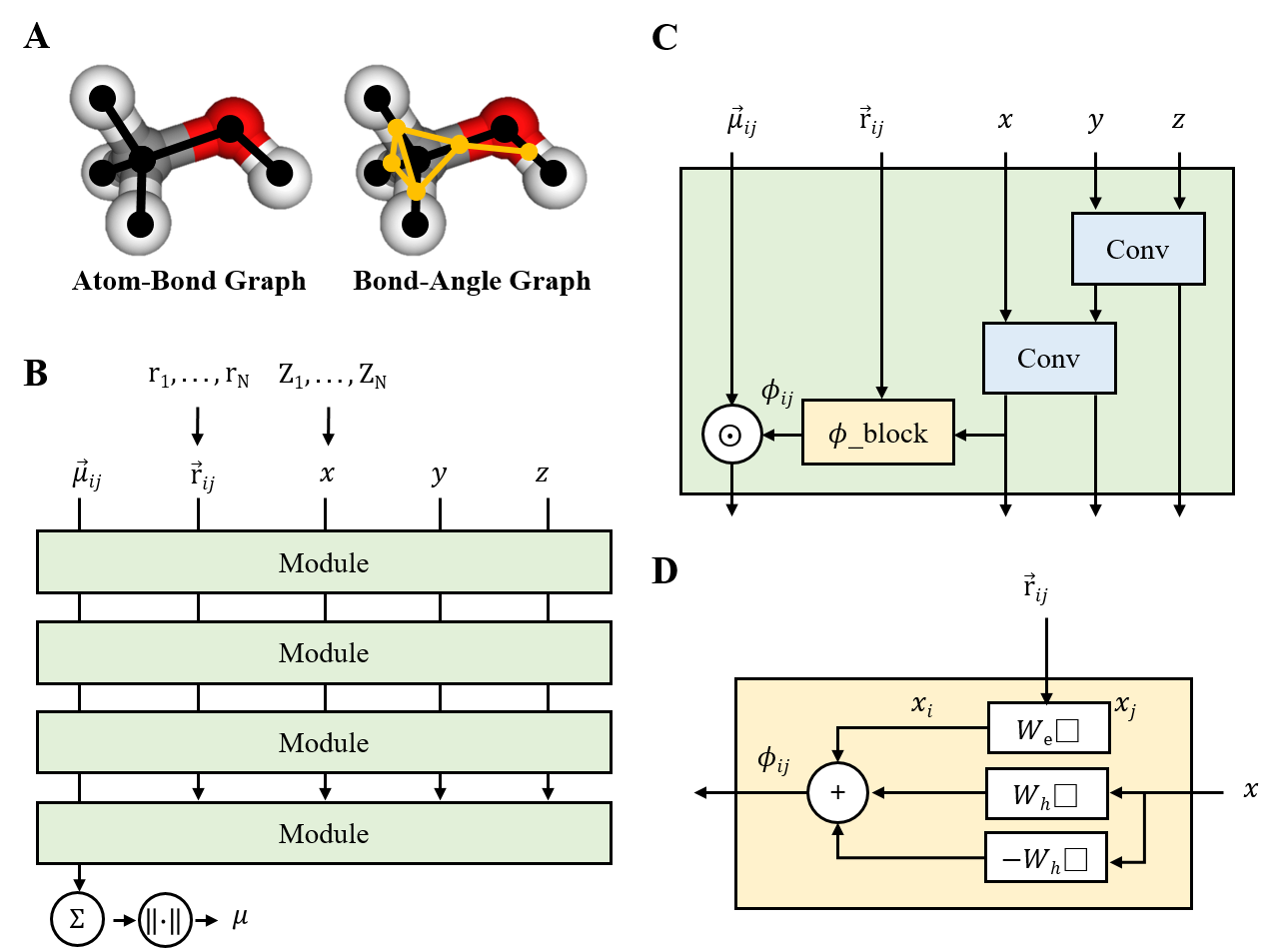}
	\caption{The overall structure of EDiGNN. Each atom, bond, and angle feature corresponds to $x$, $y$, and $z$. In the bond-angle graph, bond features are first updated through graph convolution, and in the atom-bond graph, atom features are updated using graph convolution. The atom features and displacement vectors $\textbf{r}_{ij}$ are passed through the direction-invariant module to satisfy the direction invariance of the latent vector representation $\textbf{v}_{ij}$. The dipole moment prediction $\hat{\mu}$ is calculated as the sum of the final latent vector representations.}
	\label{fig:fig2}
\end{figure}

\section{Method}
\subsection{Theory of molecular dipole moments}
The molecular dipole moment is defined by integrating the total electron charge density distribution,
\begin{equation}
    \boldsymbol{\mu } = -\int \textbf{r}\rho_e (\textbf{r})d^3 \mathbf{r} + \sum_{i}^{N}\textbf{r}_iZ_i
\end{equation}
where $\rho_e (\textbf{r})$ is the electronic charge density, $\textbf{r}_i$ is the position of the i-th nucleus, and $Z_i$ is number of protons of it. We can simplify this expression by making the approximation that partial charge $q_i$ induced by the difference between the electron charge distribution of each atom $i$ and the nuclear charge $Z_i$ has the position of that atom. The approximated dipole moment can be written as 
\cite{veit2020predicting,pereira2018machine}
\begin{equation}
    \boldsymbol{\mu } = \sum _{i}^{N} q_i\textbf{r}_i.
\end{equation}
And it can be also expanded as
\begin{equation}
    \label{eqn:dipole}
    \boldsymbol{\mu } = \sum _i^{N} q_i(\textbf{h}_i)\textbf{r}_i = \frac{1}{2(N-1)}\sum _{i,j, i\neq j}^{N} \left ( q_i(\textbf{h}_i)\textbf{r}_i + q_j(\textbf{h}_j)\textbf{r}_j \right ) = \frac{1}{2(N-1)}\sum _{i,j, i\neq j}^{N}\boldsymbol{\mu }_{ij}(\textbf{e}_{ij})
\end{equation}
where $N$ is the total number of atoms, $\boldsymbol{\mu }_{ij}$ is the local dipole moment element of an atom pair. This representation makes more sense because the position of the partial charge does not necessarily coincide with the position of the nucleus of the atom. \
The problem then becomes the determination of the $\boldsymbol{\mu }_{ij}$ that best reproduces whole dipole moment. 

\subsection{Overall model architecture}
\paragraph{Line graph framework}
Line graph framework \cite{chen2017supervised} have been a widely used framework for molecular properties predictions\cite{hsu2021efficient, kaundinya2022prediction, fang2022geometry, choudhary2021atomistic} to incorporate embeddings of bond angles into edge feature for richer representation. The line graph is a graph derived so that the edge of the original graph corresponds to a node. If two edges of the original graph share the same node, the line graph has an edge. While the nodes of an atom-bond graph correspond to atoms and its edges correspond to bonds, the nodes of a bond-angle graph as a line graph of the atom-bond graph correspond to bonds and its edges correspond to angles of bond pairs as shown in figure \ref{fig:fig2} A. Edges in the original graph and nodes in the line graph share latent representations. We use the initial atom representations from the CGCNN paper\cite{xie2018crystal}. RBF expansion is used to prepare the initial edge features in atom-bond graph and bond-angle graph. The bond angle of the triplet atoms are initially calculated by using inner product of unit vectors of bond pairs. We write atom, bond and angle representations as $x$, $y$ and $z$.
\paragraph{Graph convolution} We use graph isomorphism network (GIN) \cite{xu2018powerful} convolution for updating both node and edge features. The mathematical similarity between GNNs and the Weisfeiler-Lehman(WL) test has been noted, and it has been proven that the expressive power of a GNN cannot exceed that of 1-WL kernel. GIN is one of maximally powerful GNNs while it is simple. Since the original GIN does not incorporate edge features, we suggest GIN update as
\begin{equation}
    \mathbf{h}_i^{l+1} = MLP^l \left ( (1+ \epsilon^l) \mathbf{h}_i^l + \sum_{j \in \mathcal{N}(i)} P^l(\mathbf{e}_{ij} + \mathbf{h}_j^l) \right )
\end{equation}
where $MLP$ is multi-layer perceptron. $P^l$ is 1-layer perceptron and $\epsilon$ is learnable parameter. GIN can achieve the maximum expressive power of the WL test level under the assumption that MLP is learned as an injective function.

\paragraph{Edge embedding} 
In most GNN applications of materials field, the molecule is considered as undirected graph. In fact, a node pair has one pair of edges, from source to target and from target to source. This structure does not guarantee the symmetry of embeddings of the edge pair. If the invariance of the edge direction of the embedding is not respected, there is a risk of learning a potentially useless representation. On the other hand, if symmetry with respect to the direction of the edges of the embedding is enforced, the model representation may be restricted. To solve this, we propose a latent vector embedding that is invariant in the edge direction. We use weights with different signs for source node embeddings $\mathbf{h}_i$ and target node embeddings $\mathbf{h}_j$ to ensure symmetry of edge direction: 
\begin{equation}
    \phi_{ij} = W_h \mathbf{h}_{i} - W_h \mathbf{h}_{j} +W_r \mathbf{r}_{ij}
\end{equation}
\begin{equation}
    \mathbf{v}_{ij} = \phi_{ij} \mathbf{r}_{ij}
\end{equation}
where $W_h$ is weight matrix for node embeddings and $W_r$ is weight matrix for atomic dislocation vector. $\mathbf{v}_{ij}$ is edge direction invariant since $\phi_{ij}$ and $\mathbf{r}_{ij}$ change their sign simultaneously when edge direction is switched.
\paragraph{Readout}
To calculate dipole moments, we assumed that the predicted dipole moment as a sum of the function of latent vector embeddings as shown in equation \ref{eqn:dipole}. We applied sum-pooling to get the final predictions. This is consistent with the physical intuition that the total dipole moment is calculated as the sum of the local dipole moments.
\begin{equation}
    \hat{\mu} = \sum_{i,j} \mathbf{v}_{ij}
\end{equation}
We trained our model with the root mean square error loss function as
\begin{equation}
    \label{eqn:loss}
    L(\mu_{label}, \hat{\mu}) = \sqrt{\frac{1}{N}\sum_i^N (\mu_{label}-|\hat{\mu}|)^2}
\end{equation}

\begin{table}
	\caption{Mean Absolute Errors for prediction of dipole moments(D) of the molecules in QM9 data set. EDiGNN outperforms several models reported in the Literature by a large margin.}
	\centering
	\begin{tabular}{lrrrrrrrr}
		\toprule
		\multicolumn{5}{r}{Models}                   \\
		\cmidrule(r){2-9}
		Name     & SchNet &Cormorant & DimeNet++ &EGNN  &TFN   & ALIGNN   & PAINN&  EDiGNN\\
		\midrule
		QM9      & 0.033 & 0.038 &  0.0296 &0.029 &0.064    & 0.018  & 0.012    &  \textbf{0.010}\\
		\bottomrule
	\end{tabular}
	\label{tab:table}
\end{table}

\section{Results}

\subsection{QM9}
We test EDiGNN's performances for predicting dipole moments using the common QM9 benchmark \cite{ramakrishnan2014quantum}. It contains 134 kilo stable organic molecules consisted of C, H, O, N and F with quantum chemical properties which were calculated at the B3LYP/6-31G(2df,p) level. We use 110000 molecules for training set, 10000 molecules for validation set, and 10829 for the test set. We compare it with other state-of-the-art models such as SchNet\cite{schutt2018schnet}, Cormorant\cite{anderson2019cormorant}, DimeNet++\cite{klicpera2020fast}, EGNN\cite{satorras2021n}, Tensor field network (TFN)\cite{thomas2018tensor}, ALIGNN\cite{choudhary2021atomistic} and PAINN\cite{schutt2021equivariant}. Although the dipole moment included in the data is a scalar, the loss can be obtained by finding the norm in the final prediction as shown in Equation \ref{eqn:loss}. As shown in Table \ref{tab:table}, our methodology shows very competitive results while remaining relatively simple. We predicted the dipole moment of compounds with multiple benzene rings with a model trained in QM9 to evaluate whether the model has adequately learned the geometrical semantics of the molecule. Since the benzene ring is geometrically symmetric, all dipole moments should cancel out and be predicted to be zero. However, even the latest models fail to predict this correctly. In the case of dimenet++, since summation is used as the readout function, it is expected that the dipole moment continues to increase as the benzene ring increases. On the other hand, since ALIGNN uses mean as a readout function, it converges to a specific value when the number of carbon and hydrogen atoms is appropriately increased. The prediction results of the aforementioned two models violate the laws of physics, and it can be seen that they did not learn by reflecting the geometrical information of molecules. On the other hand, our model accurately grasps the symmetry of the benzene ring, confirming that the total dipole moment is correctly predicted to be zero.

\begin{figure}
	\centering
	\includegraphics[width=450pt]{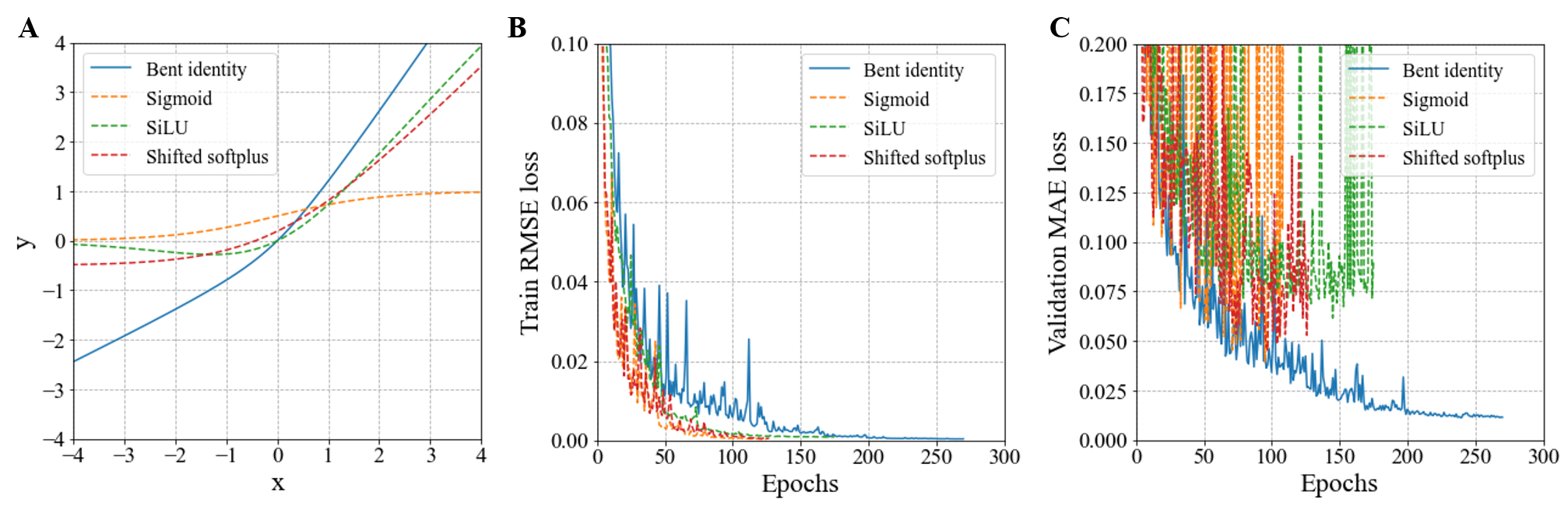}
	\caption{Activation functions used in comparative experiments (A). Training (B) and validation (C) curves with respect to activation function selection.}
	\label{fig:fig3}
\end{figure}

\subsection{Ablation studies}
\begin{figure}
	\centering
	\includegraphics[width=250pt]{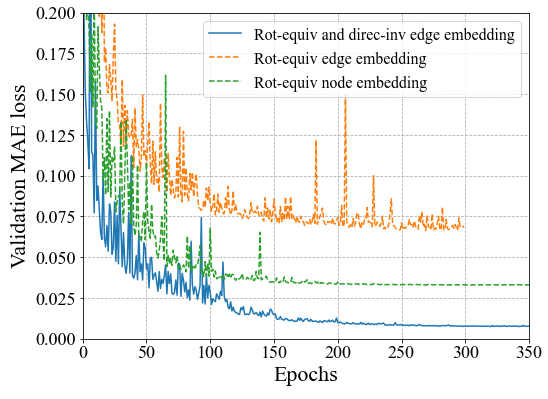}
	\caption{The dipole moment prediction expressed as the sum of the rotation invariant node representations shows a more precise result than the dipole moment expressed as the sum of the rotation invariant edge representations. Giving direction invariance to rotation invariant edge representations outperforms rotation invariant node representations.}
	\label{fig:fig4}
\end{figure}

\paragraph{Activation function}
We further investigate the performance of our model with different activation functions. SiLU, mish and shifted softplus are widely used as the activation functions of MLP in the quantum chemical property prediction task, which is a general application in the field of materials. This is to ensure the differentiability and continuity of the derivative function. Unfortunately, SiLU and mish are not injective functions. The shifted softplus activation function is one variant of softplus designed to aid convergence by allowing negative outputs. Softplus and its variants are easily squashed and numerically limited although they are injective functions. These activation functions make it difficult to satisfy injective MLP, which is the condition for securing the maximum expressive power of GIN. Bent identities, on the other hand, are injective and unbounded. 
\begin{equation}
    y = \frac{\sqrt{x^2+1}-1}{2}+x
\end{equation}
As can be seen in Figure \ref{fig:fig3} B, models using SiLU, mish, and shifted softplus as activation functions can converge the training loss, but have difficulty in generalization. The model with bent identity activation is slow to converge to the training loss, but it succeeds in generalization. 
\paragraph{Direction-invariant Edge embedding}
To test whether direction-invariant edge embeddings are the actual reason for improved performances, we compared them with rotational-equivariant edge embeddings and node embeddings. Figure \ref{fig:fig4} shows that our contributions significantly improve performance of the model. Using rotational-equivariant and direction-invariant edge embeddings decreases the error by $70\%$, which shows that this embedding provides helpful inductive bias. 

\section{Conclusion}
The efficient representation of molecular geometric information is important for predicting molecular properties. Existing work applied to molecular graphs fails to utilize molecular geometry to predict dipole moments described by vector sums, especially as is prominent in the example of aromatic compounds. To this end, we introduce a geometry-based latent vector representation to capture 3D molecular structures. This representation is designed to be invariant in the direction of the edges in a molecular graph, which is considered an undirected graph. We conducted an experiment to verify the effect of direction-invariant edge embedding and verified its superiority by comparing it with several competitive baseline models. The proposed embedding method can be useful for the related work of predicting vector properties of molecules, and we will try to apply the proposed embeddings to more molecular vector property prediction tasks in our future works.

\section{Implementation and training}
The model is implemented in Pytorch\cite{paszke2019pytorch} and deep graph library (DGL)\cite{wang2019deep}.
We train all models for 500 epochs using AdamW optimizer with normalized weight decay of $10^{-5}$.
The learning rate is reduced as half on plateau which means when a validation loss has stopped improving. The starting learning rate was $10^{-3}$. 
The final vector representations are reduced by edge-wise sum pooling and regressed to target.
Each model is trained on a single NVIDIA RTX3090 24 GB Graphics processing unit (GPU).

\bibliographystyle{naturemag}
\bibliography{references}  

\begin{thebibliography}{10}
\expandafter\ifx\csname url\endcsname\relax
  \def\url#1{\texttt{#1}}\fi
\expandafter\ifx\csname urlprefix\endcsname\relax\def\urlprefix{URL }\fi
\providecommand{\bibinfo}[2]{#2}
\providecommand{\eprint}[2][]{\url{#2}}

\bibitem{veit2020predicting}
\bibinfo{author}{Veit, M.}, \bibinfo{author}{Wilkins, D.~M.},
  \bibinfo{author}{Yang, Y.}, \bibinfo{author}{DiStasio~Jr, R.~A.} \&
  \bibinfo{author}{Ceriotti, M.}
\newblock \bibinfo{title}{Predicting molecular dipole moments by combining
  atomic partial charges and atomic dipoles}.
\newblock \emph{\bibinfo{journal}{The Journal of Chemical Physics}}
  \textbf{\bibinfo{volume}{153}}, \bibinfo{pages}{024113}
  (\bibinfo{year}{2020}).

\bibitem{pereira2018machine}
\bibinfo{author}{Pereira, F.} \& \bibinfo{author}{Aires-de Sousa, J.}
\newblock \bibinfo{title}{Machine learning for the prediction of molecular
  dipole moments obtained by density functional theory}.
\newblock \emph{\bibinfo{journal}{Journal of cheminformatics}}
  \textbf{\bibinfo{volume}{10}}, \bibinfo{pages}{1--11} (\bibinfo{year}{2018}).

\bibitem{gastegger2017machine}
\bibinfo{author}{Gastegger, M.}, \bibinfo{author}{Behler, J.} \&
  \bibinfo{author}{Marquetand, P.}
\newblock \bibinfo{title}{Machine learning molecular dynamics for the
  simulation of infrared spectra}.
\newblock \emph{\bibinfo{journal}{Chemical science}}
  \textbf{\bibinfo{volume}{8}}, \bibinfo{pages}{6924--6935}
  (\bibinfo{year}{2017}).

\bibitem{liu2020data}
\bibinfo{author}{Liu, X.}, \bibinfo{author}{Meijer, G.} \&
  \bibinfo{author}{P{\'e}rez-R{\'\i}os, J.}
\newblock \bibinfo{title}{A data-driven approach to determine dipole moments of
  diatomic molecules}.
\newblock \emph{\bibinfo{journal}{Physical Chemistry Chemical Physics}}
  \textbf{\bibinfo{volume}{22}}, \bibinfo{pages}{24191--24200}
  (\bibinfo{year}{2020}).

\bibitem{das2020dipole}
\bibinfo{author}{Das, A.} \& \bibinfo{author}{Banik, B.~K.}
\newblock \bibinfo{title}{Dipole moment in medicinal research: green and
  sustainable approach}.
\newblock In \emph{\bibinfo{booktitle}{Green Approaches in Medicinal Chemistry
  for Sustainable Drug Design}}, \bibinfo{pages}{921--964}
  (\bibinfo{publisher}{Elsevier}, \bibinfo{year}{2020}).

\bibitem{matuszek2016defining}
\bibinfo{author}{Matuszek, A.~M.} \& \bibinfo{author}{Reynisson, J.}
\newblock \bibinfo{title}{Defining known drug space using dft}.
\newblock \emph{\bibinfo{journal}{Molecular informatics}}
  \textbf{\bibinfo{volume}{35}}, \bibinfo{pages}{46--53}
  (\bibinfo{year}{2016}).

\bibitem{ioakimidis2008benchmarking}
\bibinfo{author}{Ioakimidis, L.}, \bibinfo{author}{Thoukydidis, L.},
  \bibinfo{author}{Mirza, A.}, \bibinfo{author}{Naeem, S.} \&
  \bibinfo{author}{Reynisson, J.}
\newblock \bibinfo{title}{Benchmarking the reliability of qikprop. correlation
  between experimental and predicted values}.
\newblock \emph{\bibinfo{journal}{QSAR \& Combinatorial Science}}
  \textbf{\bibinfo{volume}{27}}, \bibinfo{pages}{445--456}
  (\bibinfo{year}{2008}).

\bibitem{vo2019method}
\bibinfo{author}{Vo, M.~N.}, \bibinfo{author}{Call, M.},
  \bibinfo{author}{Kowall, C.} \& \bibinfo{author}{Johnson, J.~K.}
\newblock \bibinfo{title}{Method for predicting dipole moments of complex
  molecules for use in thermophysical property estimation}.
\newblock \emph{\bibinfo{journal}{Industrial \& Engineering Chemistry
  Research}} \textbf{\bibinfo{volume}{58}}, \bibinfo{pages}{19263--19270}
  (\bibinfo{year}{2019}).

\bibitem{nebgen2018transferable}
\bibinfo{author}{Nebgen, B.} \emph{et~al.}
\newblock \bibinfo{title}{Transferable dynamic molecular charge assignment
  using deep neural networks}.
\newblock \emph{\bibinfo{journal}{Journal of chemical theory and computation}}
  \textbf{\bibinfo{volume}{14}}, \bibinfo{pages}{4687--4698}
  (\bibinfo{year}{2018}).

\bibitem{abe2018application}
\bibinfo{author}{Abe, M.}, \bibinfo{author}{Prasannaa, V.} \&
  \bibinfo{author}{Das, B.}
\newblock \bibinfo{title}{Application of the finite-field coupled-cluster
  method to calculate molecular properties relevant to electron
  electric-dipole-moment searches}.
\newblock \emph{\bibinfo{journal}{Physical Review A}}
  \textbf{\bibinfo{volume}{97}}, \bibinfo{pages}{032515}
  (\bibinfo{year}{2018}).

\bibitem{kongsted2002dipole}
\bibinfo{author}{Kongsted, J.}, \bibinfo{author}{Osted, A.},
  \bibinfo{author}{Mikkelsen, K.~V.} \& \bibinfo{author}{Christiansen, O.}
\newblock \bibinfo{title}{Dipole and quadrupole moments of liquid water
  calculated within the coupled cluster/molecular mechanics method}.
\newblock \emph{\bibinfo{journal}{Chemical physics letters}}
  \textbf{\bibinfo{volume}{364}}, \bibinfo{pages}{379--386}
  (\bibinfo{year}{2002}).

\bibitem{leenaerts2009water}
\bibinfo{author}{Leenaerts, O.}, \bibinfo{author}{Partoens, B.} \&
  \bibinfo{author}{Peeters, F.}
\newblock \bibinfo{title}{Water on graphene: Hydrophobicity and dipole moment
  using density functional theory}.
\newblock \emph{\bibinfo{journal}{Physical Review B}}
  \textbf{\bibinfo{volume}{79}}, \bibinfo{pages}{235440}
  (\bibinfo{year}{2009}).

\bibitem{hait2018accurate}
\bibinfo{author}{Hait, D.} \& \bibinfo{author}{Head-Gordon, M.}
\newblock \bibinfo{title}{How accurate is density functional theory at
  predicting dipole moments? an assessment using a new database of 200
  benchmark values}.
\newblock \emph{\bibinfo{journal}{Journal of chemical theory and computation}}
  \textbf{\bibinfo{volume}{14}}, \bibinfo{pages}{1969--1981}
  (\bibinfo{year}{2018}).

\bibitem{faber2017prediction}
\bibinfo{author}{Faber, F.~A.} \emph{et~al.}
\newblock \bibinfo{title}{Prediction errors of molecular machine learning
  models lower than hybrid dft error}.
\newblock \emph{\bibinfo{journal}{Journal of chemical theory and computation}}
  \textbf{\bibinfo{volume}{13}}, \bibinfo{pages}{5255--5264}
  (\bibinfo{year}{2017}).

\bibitem{faber2018alchemical}
\bibinfo{author}{Faber, F.~A.}, \bibinfo{author}{Christensen, A.~S.},
  \bibinfo{author}{Huang, B.} \& \bibinfo{author}{Von~Lilienfeld, O.~A.}
\newblock \bibinfo{title}{Alchemical and structural distribution based
  representation for universal quantum machine learning}.
\newblock \emph{\bibinfo{journal}{The Journal of chemical physics}}
  \textbf{\bibinfo{volume}{148}}, \bibinfo{pages}{241717}
  (\bibinfo{year}{2018}).

\bibitem{pinheiro2020machine}
\bibinfo{author}{Pinheiro, G.~A.} \emph{et~al.}
\newblock \bibinfo{title}{Machine learning prediction of nine molecular
  properties based on the smiles representation of the qm9 quantum-chemistry
  dataset}.
\newblock \emph{\bibinfo{journal}{The Journal of Physical Chemistry A}}
  \textbf{\bibinfo{volume}{124}}, \bibinfo{pages}{9854--9866}
  (\bibinfo{year}{2020}).

\bibitem{gilmer2017neural}
\bibinfo{author}{Gilmer, J.}, \bibinfo{author}{Schoenholz, S.~S.},
  \bibinfo{author}{Riley, P.~F.}, \bibinfo{author}{Vinyals, O.} \&
  \bibinfo{author}{Dahl, G.~E.}
\newblock \bibinfo{title}{Neural message passing for quantum chemistry}.
\newblock In \emph{\bibinfo{booktitle}{International conference on machine
  learning}}, \bibinfo{pages}{1263--1272} (\bibinfo{organization}{PMLR},
  \bibinfo{year}{2017}).

\bibitem{unke2019physnet}
\bibinfo{author}{Unke, O.~T.} \& \bibinfo{author}{Meuwly, M.}
\newblock \bibinfo{title}{Physnet: A neural network for predicting energies,
  forces, dipole moments, and partial charges}.
\newblock \emph{\bibinfo{journal}{Journal of chemical theory and computation}}
  \textbf{\bibinfo{volume}{15}}, \bibinfo{pages}{3678--3693}
  (\bibinfo{year}{2019}).

\bibitem{schutt2018schnet}
\bibinfo{author}{Sch{\"u}tt, K.~T.}, \bibinfo{author}{Sauceda, H.~E.},
  \bibinfo{author}{Kindermans, P.-J.}, \bibinfo{author}{Tkatchenko, A.} \&
  \bibinfo{author}{M{\"u}ller, K.-R.}
\newblock \bibinfo{title}{Schnet--a deep learning architecture for molecules
  and materials}.
\newblock \emph{\bibinfo{journal}{The Journal of Chemical Physics}}
  \textbf{\bibinfo{volume}{148}}, \bibinfo{pages}{241722}
  (\bibinfo{year}{2018}).

\bibitem{klicpera2020directional}
\bibinfo{author}{Klicpera, J.}, \bibinfo{author}{Gro{\ss}, J.} \&
  \bibinfo{author}{G{\"u}nnemann, S.}
\newblock \bibinfo{title}{Directional message passing for molecular graphs}.
\newblock \emph{\bibinfo{journal}{arXiv preprint arXiv:2003.03123}}
  (\bibinfo{year}{2020}).

\bibitem{klicpera2020fast}
\bibinfo{author}{Klicpera, J.}, \bibinfo{author}{Giri, S.},
  \bibinfo{author}{Margraf, J.~T.} \& \bibinfo{author}{G{\"u}nnemann, S.}
\newblock \bibinfo{title}{Fast and uncertainty-aware directional message
  passing for non-equilibrium molecules}.
\newblock \emph{\bibinfo{journal}{arXiv preprint arXiv:2011.14115}}
  (\bibinfo{year}{2020}).

\bibitem{choudhary2021atomistic}
\bibinfo{author}{Choudhary, K.} \& \bibinfo{author}{DeCost, B.}
\newblock \bibinfo{title}{Atomistic line graph neural network for improved
  materials property predictions}.
\newblock \emph{\bibinfo{journal}{npj Computational Materials}}
  \textbf{\bibinfo{volume}{7}}, \bibinfo{pages}{1--8} (\bibinfo{year}{2021}).

\bibitem{schutt2021equivariant}
\bibinfo{author}{Sch{\"u}tt, K.}, \bibinfo{author}{Unke, O.} \&
  \bibinfo{author}{Gastegger, M.}
\newblock \bibinfo{title}{Equivariant message passing for the prediction of
  tensorial properties and molecular spectra}.
\newblock In \emph{\bibinfo{booktitle}{International Conference on Machine
  Learning}}, \bibinfo{pages}{9377--9388} (\bibinfo{organization}{PMLR},
  \bibinfo{year}{2021}).

\bibitem{lim2019predicting}
\bibinfo{author}{Lim, J.} \emph{et~al.}
\newblock \bibinfo{title}{Predicting drug--target interaction using a novel
  graph neural network with 3d structure-embedded graph representation}.
\newblock \emph{\bibinfo{journal}{Journal of chemical information and
  modeling}} \textbf{\bibinfo{volume}{59}}, \bibinfo{pages}{3981--3988}
  (\bibinfo{year}{2019}).

\bibitem{fout2017protein}
\bibinfo{author}{Fout, A.}, \bibinfo{author}{Byrd, J.},
  \bibinfo{author}{Shariat, B.} \& \bibinfo{author}{Ben-Hur, A.}
\newblock \bibinfo{title}{Protein interface prediction using graph
  convolutional networks}.
\newblock \emph{\bibinfo{journal}{Advances in neural information processing
  systems}} \textbf{\bibinfo{volume}{30}} (\bibinfo{year}{2017}).

\bibitem{tang2020self}
\bibinfo{author}{Tang, B.} \emph{et~al.}
\newblock \bibinfo{title}{A self-attention based message passing neural network
  for predicting molecular lipophilicity and aqueous solubility}.
\newblock \emph{\bibinfo{journal}{Journal of cheminformatics}}
  \textbf{\bibinfo{volume}{12}}, \bibinfo{pages}{1--9} (\bibinfo{year}{2020}).

\bibitem{satorras2021n}
\bibinfo{author}{Satorras, V.~G.}, \bibinfo{author}{Hoogeboom, E.} \&
  \bibinfo{author}{Welling, M.}
\newblock \bibinfo{title}{E (n) equivariant graph neural networks}.
\newblock In \emph{\bibinfo{booktitle}{International Conference on Machine
  Learning}}, \bibinfo{pages}{9323--9332} (\bibinfo{organization}{PMLR},
  \bibinfo{year}{2021}).

\bibitem{xu2018powerful}
\bibinfo{author}{Xu, K.}, \bibinfo{author}{Hu, W.}, \bibinfo{author}{Leskovec,
  J.} \& \bibinfo{author}{Jegelka, S.}
\newblock \bibinfo{title}{How powerful are graph neural networks?}
\newblock \emph{\bibinfo{journal}{arXiv preprint arXiv:1810.00826}}
  (\bibinfo{year}{2018}).

\bibitem{batzner20223}
\bibinfo{author}{Batzner, S.} \emph{et~al.}
\newblock \bibinfo{title}{E (3)-equivariant graph neural networks for
  data-efficient and accurate interatomic potentials}.
\newblock \emph{\bibinfo{journal}{Nature communications}}
  \textbf{\bibinfo{volume}{13}}, \bibinfo{pages}{1--11} (\bibinfo{year}{2022}).

\bibitem{chen2017supervised}
\bibinfo{author}{Chen, Z.}, \bibinfo{author}{Li, X.} \& \bibinfo{author}{Bruna,
  J.}
\newblock \bibinfo{title}{Supervised community detection with line graph neural
  networks}.
\newblock \emph{\bibinfo{journal}{arXiv preprint arXiv:1705.08415}}
  (\bibinfo{year}{2017}).

\bibitem{hsu2021efficient}
\bibinfo{author}{Hsu, T.} \emph{et~al.}
\newblock \bibinfo{title}{Efficient, interpretable atomistic graph neural
  network representation for angle-dependent properties and its application to
  optical spectroscopy prediction}.
\newblock \emph{\bibinfo{journal}{arXiv preprint arXiv:2109.11576}}
  (\bibinfo{year}{2021}).

\bibitem{kaundinya2022prediction}
\bibinfo{author}{Kaundinya, P.~R.}, \bibinfo{author}{Choudhary, K.} \&
  \bibinfo{author}{Kalidindi, S.~R.}
\newblock \bibinfo{title}{Prediction of the electron density of states for
  crystalline compounds with atomistic line graph neural networks (alignn)}.
\newblock \emph{\bibinfo{journal}{arXiv preprint arXiv:2201.08348}}
  (\bibinfo{year}{2022}).

\bibitem{fang2022geometry}
\bibinfo{author}{Fang, X.} \emph{et~al.}
\newblock \bibinfo{title}{Geometry-enhanced molecular representation learning
  for property prediction}.
\newblock \emph{\bibinfo{journal}{Nature Machine Intelligence}}
  \textbf{\bibinfo{volume}{4}}, \bibinfo{pages}{127--134}
  (\bibinfo{year}{2022}).

\bibitem{xie2018crystal}
\bibinfo{author}{Xie, T.} \& \bibinfo{author}{Grossman, J.~C.}
\newblock \bibinfo{title}{Crystal graph convolutional neural networks for an
  accurate and interpretable prediction of material properties}.
\newblock \emph{\bibinfo{journal}{Physical review letters}}
  \textbf{\bibinfo{volume}{120}}, \bibinfo{pages}{145301}
  (\bibinfo{year}{2018}).

\bibitem{ramakrishnan2014quantum}
\bibinfo{author}{Ramakrishnan, R.}, \bibinfo{author}{Dral, P.~O.},
  \bibinfo{author}{Rupp, M.} \& \bibinfo{author}{Von~Lilienfeld, O.~A.}
\newblock \bibinfo{title}{Quantum chemistry structures and properties of 134
  kilo molecules}.
\newblock \emph{\bibinfo{journal}{Scientific data}}
  \textbf{\bibinfo{volume}{1}}, \bibinfo{pages}{1--7} (\bibinfo{year}{2014}).

\bibitem{anderson2019cormorant}
\bibinfo{author}{Anderson, B.}, \bibinfo{author}{Hy, T.~S.} \&
  \bibinfo{author}{Kondor, R.}
\newblock \bibinfo{title}{Cormorant: Covariant molecular neural networks}.
\newblock \emph{\bibinfo{journal}{Advances in neural information processing
  systems}} \textbf{\bibinfo{volume}{32}} (\bibinfo{year}{2019}).

\bibitem{thomas2018tensor}
\bibinfo{author}{Thomas, N.} \emph{et~al.}
\newblock \bibinfo{title}{Tensor field networks: Rotation-and
  translation-equivariant neural networks for 3d point clouds}.
\newblock \emph{\bibinfo{journal}{arXiv preprint arXiv:1802.08219}}
  (\bibinfo{year}{2018}).

\bibitem{paszke2019pytorch}
\bibinfo{author}{Paszke, A.} \emph{et~al.}
\newblock \bibinfo{title}{Pytorch: An imperative style, high-performance deep
  learning library}.
\newblock \emph{\bibinfo{journal}{Advances in neural information processing
  systems}} \textbf{\bibinfo{volume}{32}} (\bibinfo{year}{2019}).

\bibitem{wang2019deep}
\bibinfo{author}{Wang, M.~Y.}
\newblock \bibinfo{title}{Deep graph library: Towards efficient and scalable
  deep learning on graphs}.
\newblock In \emph{\bibinfo{booktitle}{ICLR workshop on representation learning
  on graphs and manifolds}} (\bibinfo{year}{2019}).

\end{thebibliography}

\end{document}